\documentclass{article}
\usepackage{spconf,amsmath,graphicx}
\usepackage{amssymb}
\usepackage{tipa}
\usepackage{enumitem}
\setlist{nolistsep}
\usepackage{hyperref}

\title{Phone-to-audio alignment without text: A Semi-supervised Approach} 

\name{Jian Zhu$^{\star}$ \qquad  Cong Zhang$^{\dagger}$ \qquad David Jurgens$^{\ddagger}$}
\address{$^{\star}$Department of Linguistics, University of Michigan, Ann Arbor, USA \\
$^{\dagger}$Center for Language Studies, Radboud University, Nijmegen, Netherlands \\
$^{\ddagger}$School of Information, University of Michigan, Ann Arbor, USA \\
\texttt{lingjzhu@umich.edu, cong.zhang@ru.nl, jurgens@umich.edu}}

\begin{document}
%
\maketitle
\begin{abstract}
The task of phone-to-audio alignment has many applications in speech research. Here we introduce two Wav2Vec2-based models for both text-dependent and text-independent phone-to-audio alignment. The proposed Wav2Vec2-FS, a semi-supervised model, directly learns phone-to-audio alignment through contrastive learning and a forward sum loss, and can be coupled with a pretrained phone recognizer to achieve text-independent alignment. The other model, Wav2Vec2-FC, is a frame classification model trained on forced aligned labels that can both perform forced alignment and text-independent segmentation. Evaluation results suggest that both proposed methods, even when transcriptions are not available, generate highly close results to existing forced alignment tools. Our work presents a neural pipeline of fully automated phone-to-audio alignment. Code and pretrained models are available at \url{https://github.com/lingjzhu/charsiu}.

\end{abstract}
\section{Introduction}
\label{sec:intro}

Aligning phones to audio has been a fundamental task in speech and phonetic research, as many speech analyses depend on knowing the exact timing of phones. 
There are currently a few existing tools for performing forced alignment over phone sequences \cite{mcauliffe2017montreal,gorman2011prosodylab,rosenfelder2011fave,kisler2012signal,gentle}. 
The majority of them are based on the classic HMM system built upon Kaldi \cite{povey2011kaldi} or the HTK toolkit \cite{young2002htk}. Though neural ASR systems have shown better performance than HMM systems \cite{NEURIPS2020_92d1e1eb}, phone-to-audio alignment has not benefited a lot from the predictive power of neural networks. One reason is that model ASR has increasingly shifted towards end-to-end training using loss functions like CTC \cite{graves2006connectionist} that disregards precise frame alignment.  
Only a few works explored using neural networks to perform segmentation of sentences \cite{kurzinger2020ctc} and phones \cite{Kelley2018,schulze2020joint,teytaut21_interspeech}. These works demonstrate great potentials for neural forced alignment, but they still required text transcriptions. Compared to forced alignment, text-independent alignment has received much less attention. While a few unsupervised or self-supervised phone segmentation methods have been proposed \cite{Kreuk2020,kamper2020towards,bhati2021segmental,teytaut21_interspeech}, most of them (except \cite{teytaut21_interspeech}) only output phone boundaries but do not jointly predict \textit{both} boundaries and phones, making them less practical in textless alignment. Therefore, we also developed a method to perform alignments directly on audio, as transcriptions might not always be available in many practical applications.

In this study, we present two neural networks for both text-dependent and text-independent phone-to-audio alignment. Our contribution can be summarized as follows. First, we present a semi-supervised model to perform forced alignment, but can also be combined with a phone recognizer for text-independent alignment. It achieves comparable results with four existing forced alignment tools. Secondly, we have also tested a frame classification model for text-independent alignment, which achieved close performance to text-dependent models. 
Extensive analyses show that the proposed methods still maintain good performance in different settings.
In contrast to currently available tools, our work provides speech and language researchers with a strong deep learning based tool to obtain phone segmentations even when textual transcriptions are not available.

\section{Neural forced alignment}

Given a speech signal $X^{S}\in\mathbb{R}^{1\times T_{raw}}$ and a sequence of phonetic symbols $Y^{P}\in\mathbb{R}^{1\times N}$, some neural networks are used to encode them into hidden representations $\Phi$, such that $\boldsymbol{Y} = \Phi_S(Y_{P})$ and $\boldsymbol{X} = \Phi_P(X_{S})$, where
$\boldsymbol{X}\in \mathbb{R}^{K\times T}=[\boldsymbol{x}_1,\boldsymbol{x}_2,\dots,\boldsymbol{x}_t]$ and $\boldsymbol{Y} \in \mathbb{R}^{K\times N} = [\boldsymbol{y}_1,\boldsymbol{y}_2,\dots,\boldsymbol{y}_n]$, $T\geq N$. We seek to learn a transformation $f$ to encode these features into the monotonic alignment $S=f(\boldsymbol{X},\boldsymbol{Y})$, so that it aligns each phonetic symbol $y^P_m$ to a subset of consecutive acoustic frames $\boldsymbol{x}_i$: $S$ = $[\boldsymbol{x}_i=y^P_m,\boldsymbol{x}_j=y^P_m,\dots,\boldsymbol{x}_k=y^P_n]$. If transcriptions are not available, phone labels $\hat{Y}^P$ can be estimated with a phone recognizer from audio.

The proposed neural forced alignment model learns the phone-to-audio alignment through the self-supervised task of reconstructing the quantized embeddings of original speech with both heavily masked speech representations and phonemic information \cite{schulze2020joint}. This could be implemented as the same pretraining task of Wav2Vec2 \cite{NEURIPS2020_92d1e1eb}. Meanwhile, we relied on the forward sum loss to guide the attention matrix to be monotonic alignments \cite{shih2021rad,badlani2021one}.

\subsection{Encoders}
We used the pretrained \texttt{wav2vec2-base} model \cite{NEURIPS2020_92d1e1eb} with a convolutional layer on top as the speech encoder $\Phi_{S}$. As the original frame rate of Wav2Vec2 (about 20ms per frame) could be too coarse to derive precise timing, we increased the frame rate by two methods, either by reducing the stride of the last convolutional layer in its feature extractor from 2 to 1 (denoted as \texttt{-10ms}), or by upsampling the raw speech signal to 32kHz (denoted as \texttt{-32k}). This effectively doubled the frame rate to 98Hz (about 10ms per frame), while allowing the model to load all the pretrained parameters of \texttt{wav2vec2-base}. 

The phone encoder $\Phi_{T}$ is a reduced version of the BERT model \cite{devlin-etal-2019-bert} with a convolutional layer on top. It has 4 hidden layers, each of which has 12 self-attention heads of 32 hidden dimensions. The model was pretrained with masked language modeling \cite{devlin-etal-2019-bert} on phone sequences generated from the whole Book Corpus \cite{Zhu_2015_ICCV}, which amounts to more than 2 billion of phone tokens.

\subsection{Loss function}

The model learns a matrix $\boldsymbol{A}\in\mathbb{R}^{N\times T}$ that aligns the phone sequence $\boldsymbol{Y}$ and the masked speech representation $\boldsymbol{\hat{X}}$. 

\begin{equation}
    \boldsymbol{D}_{ij}=f_y(\boldsymbol{y}_i)^Tf_x(\boldsymbol{\hat{x}}_j)
\end{equation}
\begin{equation}
    \boldsymbol{A} = softmax(\boldsymbol{D},\texttt{dim=0})
\end{equation}
\begin{equation}
    \boldsymbol{H} = concatenate[\boldsymbol{\hat{X}},\boldsymbol{Y}\boldsymbol{A}]
\end{equation}

where $f_y$ and $f_x$ are two dense layers. $\boldsymbol{D}\in\mathbb{R}^{N\times T}$ is the (unnormalized) similarities between $\boldsymbol{Y}$ and $\boldsymbol{X}$. For the hidden states $\boldsymbol{H}\in \mathbb{R}^{2K\times T}=[\boldsymbol{h}_1,\dots,\boldsymbol{h}_T]$ at each time step, the model maximizes the similarity between each $\boldsymbol{h}_t$ and the quantized version of the acoustic feature at the same time step. The quantized embeddings can be selected from a codebook using Gumbel softmax parameterized by a neural network, as in the original Wav2Vec2 \cite{NEURIPS2020_92d1e1eb}. Let $\boldsymbol{Q}_t$ represents the set consisting of a true quantized embedding and $n=50$ negative samples at step t. The loss function is formulated as follows.
\begin{equation}
    \mathcal{L}_{m} = \frac{1}{T} \sum_{t=1}^{T} -\log\frac{\exp{(sim(\boldsymbol{h}_t,\boldsymbol{q}_t)/\kappa)}}{\sum_{\boldsymbol{q}_k\sim\boldsymbol{Q}_t}\exp(sim(\boldsymbol{h}_t,\boldsymbol{q}_k)/\kappa)}
\end{equation}
The masking is performed with spectral augmentation with both time and feature masking \cite{park19e_interspeech}. If the time masking probability is too low, the model might simply ignore the phone information. However, if the time masking probability is too high, it will create a gap between training and inference inputs, as the input audio for inference is usually not corrupted. We sampled the time-masking probability $p/100$ for each batch from a discrete uniform distribution of $[p_l,p_h]$. The original weights of the Wav2Vec2 quantizer were kept fixed throughout training.

Yet the attention matrices might not necessarily align acoustic frames and phone tokens. Therefore we added the forward-sum loss used in HMM systems to constrain the attention matrix to be monotonic and diagonal \cite{shih2021rad,badlani2021one}. 
\begin{equation}
    \mathcal{L}_{FS} = -\sum_{\boldsymbol{X},\boldsymbol{Y}\in S}\log P(\boldsymbol{Y}|\boldsymbol{X})
\end{equation}
where $S$ is the optimal alignment. All probability quantities are estimated in the attention matrix $\boldsymbol{A}$. This loss function can be implemented using the off-the-shelf PyTorch CTC loss by setting the probability of the blank symbol to an impossibly low value. We adapted the implementation from \cite{shih2021rad,badlani2021one}. So the final loss is a weighted sum of the two loss functions $\mathcal{L}=\mathcal{L}_m + \lambda\mathcal{L}_{FS}$, where $\lambda$ was set to 1.

The alignment is notoriously hard to train, since early deviations from the ground truth only get reinforced in subsequent iterations \cite{zeyer17_interspeech}. 
Techniques such as guided attention loss \cite{tachibana2018efficiently} or diagonal prior \cite{shih2021rad,badlani2021one} or external supervision with available alignments \cite{ren2020fastspeech} have been used to enable the formation of text-to-speech alignment. To facilitate alignment learning, we train the model incrementally in the curriculum learning paradigm \cite{10.1145/1553374.1553380} on increasingly longer audio samples. Like noisy student training \cite{park2020improved} and the iterative training in Kaldi \cite{povey2011kaldi}, a series of models are generated in the process. 

\begin{enumerate}
    \item Initialize model $M$ with pretrained weights $M_0$. Set $M=M_0$. 
    \item Partition the length-sorted data $C$ into $k$ different chunks $C_1,C_2,\dots,C_k$ by duration, such that each chunk contains audio longer than the previous chunk. Then train $M$ on each chunk sequentially with a frame shift of 20ms. 
    \item Upsample the frame shift of $M$ to 10ms. Retrain $M$ using the same data and procedure as in 2. 
    \item (\textit{Optional}) Train a frame classification model  $M^\prime$ from scratch with alignments generated from the last step. Set $M=M^\prime$ and go back to 2.
\end{enumerate}

Reinitialization with a frame classification model speeds up the convergence as it has been noted that a prior is essential for obtaining good alignment \cite{zeyer17_interspeech}. As the learned attention alignments are sharp, the decoding can be performed simply using the \texttt{argmax} function without resorting to the Viterbi decoding. This forced alignment is referred to as \texttt{W2V2-FS}.

\section{Frame classification model}
For text-independent alignment, we fine-tuned a Wav2Vec2 model to perform frame-wise phone classification (referred to as \texttt{W2V2-FC-Libris}). The training labels were alignments of the 960h \textit{Librispeech} \cite{panayotov2015librispeech} obtained through MFA used in \cite{lugosch19_interspeech}. The alignments were discretized into 10ms or 20ms frames and the model was trained to classify each frame into a phone category using the cross-entropy loss. During inference, the phone alignment was acquired through model predictions directly with audio, but forced alignment was also performed using a Dynamic Time Warping (DTW) algorithm based on an output probability matrix and its transcription. We also trained additional frame classification models (\texttt{W2V2-FC}) on the alignments generated from the semi-supervised method.

\section{Experiments}

\begin{table}[]
\centering
\scriptsize
\caption{Evaluation results of text-dependent alignment}

\begin{tabular}{lccccc}
\hline
Model & P & R & F1 & R-val & Overlap \\ \hline
\texttt{FAVE}       & 0.57  & 0.59  & 0.58   &  0.64   &    74.3\%       \\
\texttt{MFA-Libris}       & 0.61  & 0.61  &  0.61  &    0.67   & 73.5\%      \\
\texttt{MFA}       & 0.62  & 0.63  &  0.63  &   0.68 & 75.0\%     \\
\texttt{Gentle}       & 0.49  & 0.46 & 0.48  &  0.56    &  67.7\%     \\ 
\texttt{WebMAUS}       & \textbf{0.70}  & \textbf{0.70}  & \textbf{0.70}   &  \textbf{0.75}       &   78.8\%   \\ \hline
\texttt{W2V2-FC-20ms-Libris}  & 0.49 & 0.47 &  0.48  &  0.56  &   73.8\%  \\
\texttt{W2V2-FC-10ms-Libris}  & 0.57  &  0.54 & 0.55   &   0.62  &    76.4\%  \\
\texttt{W2V2-FC-32k-Libris}  & 0.66  & 0.63  & 0.64   & 0.69  & 79.3\%     \\\hline
\texttt{W2V2-FS-20ms} & 0.47 & 0.49 & 0.48 & 0.55 & 71.6\% \\
\texttt{W2V2-FS-10ms} & 0.68 & 0.68 & 0.68 & 0.73 & \textbf{80.4}\% \\
\texttt{W2V2-FS-32k} & 0.63 & 0.65 & 0.64 & 0.69 & 79.3\% \\\hline
\multicolumn{6}{l}{\textit{Pretrained G2P converter} }\\\hline
\texttt{W2V2-FS-20ms} & 0.40 & 0.42 & 0.41 & 0.49 & 65.1\% \\
\texttt{W2V2-FS-10ms} & 0.56 & \textbf{0.58} & 0.57 & 0.63 & 72.5 \\
\texttt{W2V2-FC-32k-Libris} & \textbf{0.58} & 0.57 & 0.58 & \textbf{0.64} & \textbf{73.0\%} \\\hline
\multicolumn{6}{l}{\textit{Phone set adaptation} (\texttt{TIMIT-61}) } \\\hline
\texttt{W2V2-FS-20ms} & 0.49 & 0.53 & 0.51 & 0.57 & 70.5\% \\
\texttt{W2V2-FS-10ms} & \textbf{0.66} & \textbf{0.70} & \textbf{0.68} & \textbf{0.72} & \textbf{79.7\%} \\ \hline
\end{tabular}
\label{tab:de}
\end{table}

For the English dataset, we used the Librispeech \cite{panayotov2015librispeech} the 1600h \textit{Common Voice} 6.1 \cite{ardila2020common}. The TIMIT dataset with human annotations \cite{garofolo1993darpa} was used for evaluation. The original TIMIT 61 phone annotations were collapsed into the 39 CMU phone set. The flap [\textipa{R}] \texttt{DX} did not have a one-to-one mapping, so we kept all occurrences. While this slightly degraded the overall accuracy, this was the same for all forced alignment methods under comparison except \texttt{WebMAUS}, which used more phones than those in the CMU phone set. All data and their train/test partitions were based on the HuggingFace \texttt{datasets} package \cite{lhoest2021datasets}. All textual transcriptions were converted into phone sequences using an open source grapheme-to-phoneme (G2P) converter \cite{g2pE2019}.

In our experiment, we ran four iterations of training. We set the time masking probability to $[0.05,0.2]$ for the 20ms models and $[0.05,0.4]$ for 10ms models. The first two iterations were trained on Common Voice. We partitioned the Common Voice dataset into three sets $\{C1,C2,C3\}$, $C1$ containing sentences below 3 seconds, $C2$ with sentences between 3-5 seconds and $C3$ consisting of sentences up to 10 seconds. The third iteration was trained on Librispeech and in the fourth iteration, the model was fine-tuned on the TIMIT train set. The model generally 
converged within a thousand training steps so training on full dataset was not needed.
Annotated boundary information was never used during training. Evaluation was performed on the TIMIT test set, which does not overlap with any of the training data. The proposed model was coded in \texttt{pytorch} based on the Wav2Vec2 implementations in the \texttt{transformers} package.
By default, we used an effective batch size of 32 and the Adam optimizer with a learning rate of 1e-6 and rate decay of 1e-6. The training was done distributedly on three GPUs with 12 GB of memory.

We compared our method with four publicly available forced aligners: \texttt{Montreal Forced Aligner (MFA)} \cite{mcauliffe2017montreal}, \texttt{Penn Forced Aligner (FAVE)} \cite{rosenfelder2011fave}, \texttt{WebMAUS} \cite{kisler2012signal} and \texttt{Gentle} \cite{gentle}. Specifically for \texttt{MFA}, we tested its performance both using a pretrained Librispeech model and performing forced alignment from scratch on TIMIT.

\section{Evaluation}

\begin{table}[]
\centering
\scriptsize
\caption{Evaluation results of text-independent alignment}

\begin{tabular}{lccccc}
\hline
Model & P & R & F1 & R-val & Overlap \\ \hline
\texttt{W2V2-CTC-10ms}   & 0.31  & 0.29  & 0.30   &  0.42       &   43.9\%   \\ 
\texttt{W2V2-CTC-20ms}   & 0.31  & 0.30 & 0.31   &  0.42      &   46.6\%  \\ \hline
\multicolumn{6}{l}{\textit{Phone recognition} + \texttt{W2V2-FS}} \\\hline
\texttt{W2V2-FS-20ms} & 0.40 & 0.42 & 0.41 & 0.48 &  64.2\% \\
\texttt{W2V2-FS-10ms} & 0.56 & 0.58 & 0.57 & 0.63 & 71.5\% \\
\texttt{W2V2-FC-32k-Libris}  &  0.57 & 0.57 & 0.57 & 0.64 & 72.2\% \\ \hline
\textit{Direct inference } \\\hline
\texttt{W2V2-FC-20ms-Libris}  & 0.57  &  0.59 & 0.58   &   0.63  &    72.7\%  \\
\texttt{W2V2-FC-10ms-Libris}  & 0.55  &  0.58 & 0.56   &   0.62  &    72.5\%  \\
\texttt{W2V2-FC-32k-Libris}  & \textbf{0.60}  & \textbf{0.63}  & \textbf{0.61}   &  \textbf{0.66}  &   \textbf{74.3\%}   \\ \hline
\end{tabular}
\label{tab:inde}
\end{table}

We evaluated the forced alignment results with precision, recall, F1, and R-value 
\cite{Kreuk2020}. For each predicted phone boundary, if the timing was within tolerance $\tau$ and the predicted phone matches, it was considered a hit. As each boundary marked the onset and the offset of consecutive phones, we only evaluated the phone onsets with a tolerance of 20ms. In addition, we also measured the percentage of correctly predicted frame labels at a 10ms time scale. 

In Table~\ref{tab:de}, all methods can perform forced alignment with decent accuracy, though \texttt{Gentle} seems to lag slightly behind other methods.
The proposed forced alignment method also shows comparable performance, comparable to all existing forced alignment methods, if not better. In general, increasing the frameshift from 20ms to 10ms improved the alignment accuracy. However, phones in continuous speech often undergo a variety of phonetic processes such as deletion and insertion, resulting in pronunciation variants deviating from dictionary pronunciations. Yet most G2P converters can only provide dictionary pronunciations, which may lead to decreased performance. To shed light on this issue, we tested the influence of a pretrained G2P converter.
In Table~\ref{tab:de}, it is clear that using the dictionary pronunciations decreases the alignment performance by a significant margin, yet the performance is still close to current forced aligners.

As different speech varieties may have varying phone inventories (like American English vs. Received Pronunciation), it would be desirable for the pretrained \texttt{W2V2-FS} model to adapt to different phone inventories. To test this, we can simply fine-tune the model after modifying the phone embeddings of the text encoder. For TIMIT, we expanded the existing 39 phone embeddings of the original \texttt{W2V2-FS} model to include the full TIMIT 61 phone set while keeping the rest of the pretrained weights intact, and fine-tuned the \texttt{W2V2-FS} model. The model converged quickly after a few hundred iterations and achieved comparable performance as the original model (see last two rows in Table~\ref{tab:de}). Yet \texttt{W2V2-FC} models might be less flexible in this regard, as it requires pre-existing alignments to train.

For text-independent alignment, we trained a phone recognizer on both Librispeech and Common Voice by fine-tuning \texttt{wav2vec2-base} using the CTC loss. This phone recognizer was used to derive phone transcriptions first before performing forced alignment with \texttt{W2V2-FS}. Using a phone recognizer works reasonably well (Table~\ref{tab:inde}), as the alignment is only 1\% worse than using a pretrained G2P converter.  As expected, the alignments generated by the CTC loss alone deviated significantly from the ground truth, since the CTC loss does not encourage time-synchronous alignment. In contrast, \texttt{W2V2-FC} models achieved better performance (Table~\ref{tab:inde}) than other text independent models, highly comparable to that of text-dependent alignment methods. This proves that the frame classification model used in HMM systems are still robust with neural networks. 
Overall, these encouraging results suggest that both text independent models are practical means to derive phone-to-audio alignments.

\begin{figure}
    \centering
    \includegraphics[width=0.9\linewidth]{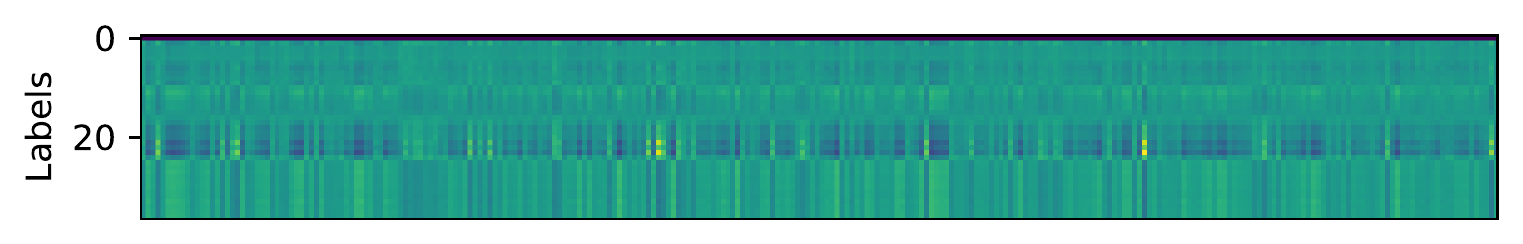}
    
    \vspace{-3mm} 
    
    \includegraphics[width=0.9\linewidth]{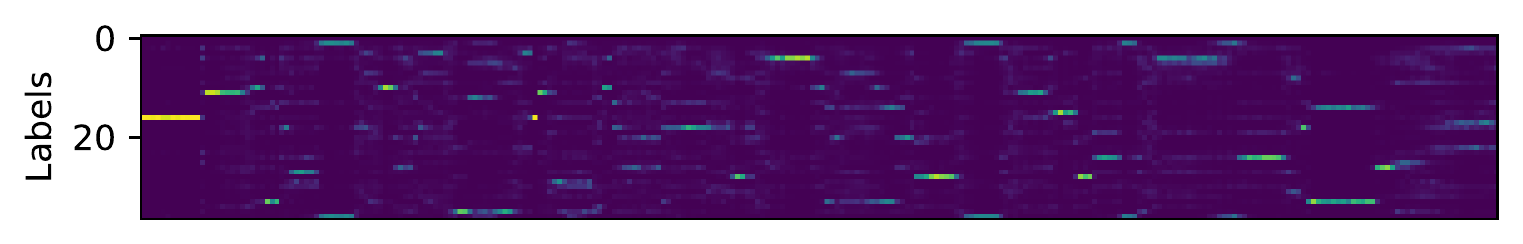}
    
    \vspace{-3mm}
    
    \includegraphics[width=0.9\linewidth]{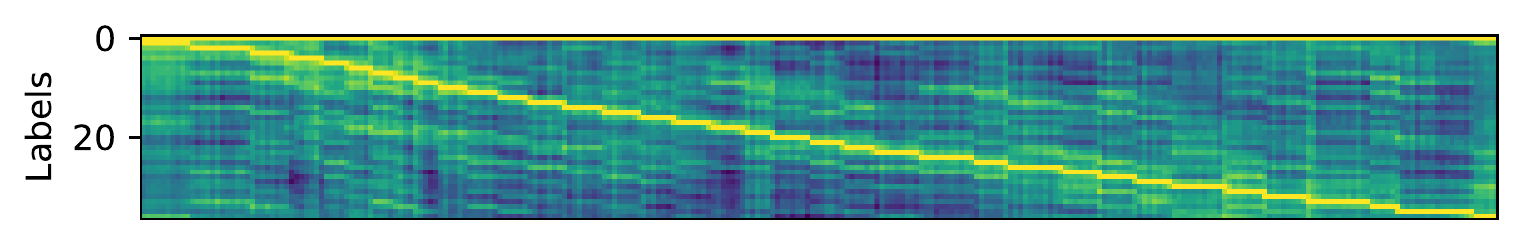}
    
    \vspace{-3mm}
    
    \includegraphics[width=0.9\linewidth]{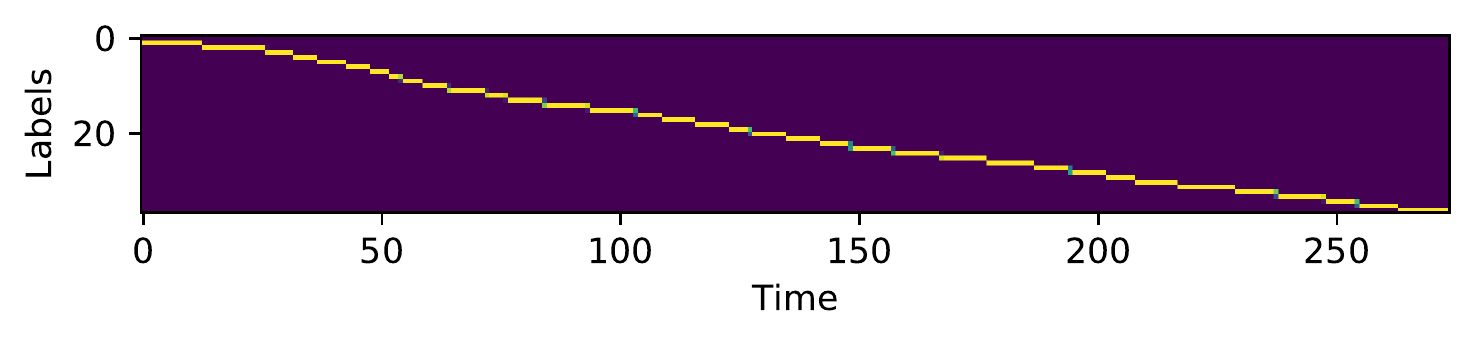}
        
    \vspace{-3mm}

    \caption{Sample alignments. From top to bottom: 1) No curriculum training; 2) No $\mathcal{L}_{FS}$; 3) No $\mathcal{L}_m$; 4) Full model.}
    \label{fig:my_label}
\end{figure}

\begin{table}[]
\centering
\scriptsize
\caption{Evaluation of \texttt{W2V2} models from different iterations}
\vspace{1mm}
\begin{tabular}{lccccc}
\hline
Model & Iteration &  Training Data & R-val & Overlap \\ \hline
\texttt{FS-10ms} & 1  & Common Voice & 0.41 & 60.7\% \\
\texttt{FC-10ms(w/o text)} & 1  & Common Voice & 0.37  & 55.8\% \\
\texttt{FS-10ms} & 2  & Common Voice & 0.51 & 68.1\% \\
\texttt{FC-10ms(w/o text)} & 2  & Common Voice & 0.53 & 66.2\% \\
\texttt{FS-10ms} & 3  & Librispeech & 0.60 & 74.1\% \\
\texttt{FC-10ms(w/o text)} & 3  & Librispeech & 0.52 & 68.1\% \\
\texttt{FC-10ms(w/ text)} & 3  & Librispeech & \textbf{0.63} & \textbf{75.3\%} \\\hline
\end{tabular}
\label{tab:iterations}
\end{table}

When the errors of G2P are taken into account, our forced alignment methods might not possess a strong advantage over existing tools. The discrepancy could be accounted for by two causes. First, silence poses challenges to forced alignment as it is not encoded in the transcription. Without preprocessing silence, phones are often misaligned when a long silent interval is present. Secondly, we did not explicitly model pronunciation variants. However, HMM-based systems \cite{povey2011kaldi} often have built-in methods to alleviate these two issues. This will be the next step for our research.

\section{Ablation analysis}
We performed ablation analysis to examine the effectiveness of the \texttt{W2V2-FS} model. As shown in Fig~\ref{fig:my_label}, both the contrastive loss and the forward-sum loss are necessary to facilitate the learning of sharp attention alignment. If the audio and the transcription were aligned using only the forward sum loss, the learned alignment was noisy. If either the curriculum learning or the contrastive loss is removed, the model generally fails to converge to a diagonal alignment pattern. 

Multiple iterations of training are generally needed to boost performance (see Table~\ref{tab:iterations}), as initial alignments tend to be suboptimal, especially for noisy datasets like Common Voice. During training, we found that the forced alignment model was \textit{very} sensitive to initialization, and poor alignment was reinforced but not corrected in subsequent training. So using a frame classification to provide a good prior estimation was important.
Audio quality had a large impact on the quality of alignment. Training on TIMIT or Librispeech-clean from scratch ended up with better alignments, faster convergence rates, and fewer iterations than training on the crowd-sourced Common Voice. Table~\ref{tab:iterations} illustrates the variations of model performance when noisy data and audio domain mismatch are both present, though the performance gaps can be narrowed through fine-tuning.

\section{Discussions and Conclusions}
In this study, we present two deep learning-based methods to align audio to phones and achieve comparable performance against several existing forced alignment tools. These models can be combined as a pipeline to bootstrap phone labels from naturalistic audio data in the wild, which are massive but remain under-exploited. This will have implications for speech corpus creation, phonetic research, and interdisciplinary studies. Given the good performance of $\texttt{W2V2-FC-Libris}$, our method can also be combined with MFA\cite{mcauliffe2017montreal}  to train a large-scale text-independent frame classification model to get rid of text in inference. The W2V2 models 
can be accelerated by harnessing the computing power of GPU. For example, in the inference mode, our model aligned the 1600 hours Common Voice dataset in about 10 hours on a 2080Ti GPU without batching. While only tested on phone segmentation, the model also performed well on related tasks such as word or sentence segmentation. 
In the future, we will continue to improve the alignment accuracy and extend it to different languages.

\vfill\pagebreak
\bibliographystyle{IEEEbib}
\bibliography{refs}

\end{document}